\documentclass[11pt]{article}

\usepackage{graphicx} 
\usepackage{subcaption}
\usepackage{float}
\usepackage[ruled,lined,linesnumbered,commentsnumbered]{algorithm2e}  
\usepackage[most]{tcolorbox}
\usepackage{lipsum}
\usepackage[letterpaper, top=1in, bottom=1in, left=1in, right=1in]{geometry}
\usepackage{minitoc}
\usepackage{tablefootnote}
\usepackage{authblk}
\usepackage{xcolor}
\usepackage{minitoc}
\usepackage[utf8]{inputenc} 
\usepackage[T1]{fontenc}    
\usepackage{hyperref}       
\usepackage{url}            
\usepackage{booktabs}       
\usepackage{amsfonts}       
\usepackage{nicefrac}       
\usepackage{microtype}      
\usepackage{xcolor}  
\usepackage{wrapfig} 

\usepackage[final]{acl}
\usepackage{titletoc}
\usepackage{times}
\usepackage{latexsym}

\usepackage[T1]{fontenc}

\usepackage[utf8]{inputenc}

\usepackage{microtype}

\usepackage{hyperref}  
\usepackage{xcolor}   

\hypersetup{
    colorlinks=true,  
    linkcolor=red,       
    urlcolor= blue
}

\usepackage{inconsolata}

\usepackage{graphicx}

\usepackage{times}
\usepackage{latexsym}

\usepackage[T1]{fontenc}

\usepackage[utf8]{inputenc}

\usepackage{microtype}

\usepackage{inconsolata}

\usepackage{graphicx}

\title{The \textit{Missing vs. Unused} Knowledge Hypothesis for Language Model Bottlenecks in Patent Understanding}

\author{
  Siyang Wu,
  Honglin Bao,
  Nadav Kunievsky,
  James A. Evans
\\
  University of Chicago
\\
\\
\\
  \small{\{siyangwu, honglinbao, nadavkunievsky, jevans\}@uchicago.edu}
}

\begin{document}

\maketitle

\begin{abstract}
While large language models (LLMs) excel at factual recall, the real challenge lies in knowledge application. A gap persists between their ability to answer complex questions and their effectiveness in performing tasks that require that knowledge. We investigate this gap using a patent classification problem that requires deep conceptual understanding to distinguish semantically similar but objectively different patents written in dense, strategic technical language. We find that LLMs often struggle with this distinction. To diagnose the source of these failures, we introduce a framework that decomposes model errors into two categories: \emph{missing knowledge and unused knowledge}. Our method prompts models to generate clarifying questions and compares three settings -- raw performance, self-answered questions that activate internal knowledge, and externally provided answers that supply missing knowledge (if any). We show that most errors stem from failures to deploy existing knowledge rather than from true knowledge gaps. We also examine how models differ in constructing task-specific question–answer databases. Smaller models tend to generate simpler questions that they, and other models, can retrieve and use effectively, whereas larger models produce more complex questions that are less effective, suggesting complementary strengths across model scales. Together, our findings highlight that shifting evaluation from static fact recall to dynamic knowledge application offers a more informative view of model capabilities.

\end{abstract}

\section{Introduction}

Large Language Models (LLMs) are increasingly viewed not merely as passive text generators, but as systems that can reason about and engage with complex realities. Yet despite their impressive capabilities, LLMs often struggle with tasks that hinge on subtle conceptual distinctions \citep{asthana2024evaluating, havlik2024meaning, pavlick2023symbols}. We hypothesize that this difficulty reflects not a lack of knowledge, but limits in how knowledge is represented and accessed: much of what LLMs encode remains latent, poorly organized, and difficult to retrieve in knowledge-intensive settings \citep{pan2025understanding, hoang2023compressed}. This creates a fundamental tension: although LLMs can appear to “know everything,” their knowledge is stored in a highly compressed form that can hinder activation, recall, and information-quality assessment \citep{hoang2024simple, hoang2023compressed, jaiswal2023compressing}. More broadly, the gap between the knowledge an LLM contains and what it reliably deploys at inference time is central to evaluating its competence. In humans, knowledge that is learned is typically accessible for reasoning and problem-solving; in LLMs, by contrast, having information in parameters does not guarantee its availability when needed. Prior work has examined this gap using narrowly scoped probes, such as math word problems or factual recall, to test what models do or do not use \citep{asthana2024evaluating, havlik2024meaning, pavlick2023symbols}, but such tasks often miss the holistic, integrated nature of conceptual understanding.

We refer to these two forms of knowledge as \emph{lay in knowledge}, knowledge the model can produce when directly prompted, and \emph{working knowledge}, the knowledge it actually brings to bear in the course of solving a task. 

To quantify the contribution of each type of knowledge to task performance, we design a conceptual understanding test that requires deploying knowledge dynamically rather than merely stating it. Understanding is a core cognitive capacity underlying most knowledge-intensive behavior: it goes beyond recalling facts or executing procedures, and instead involves grasping structural links and hierarchies that support transfer and generalization. Without some levels of understanding, both humans and LLMs may succeed on familiar patterns yet fail under distributional shift, novel recombinations of known components, or adversarially phrased problems. We therefore treat understanding not as an optional “nice-to-have,” but as a prerequisite for turning stored information into usable working knowledge. Our testbed is patent understanding, where success depends on fine-grained access to mechanisms, purposes, and constraints.\footnote{Section \ref{sec:whatIsUnderstanding} in the appendix offers a more detailed discussion of the link between understanding and pairwise differentiation.}

Drawing from cognitive science and philosophy, we posit that a key, observable facet of understanding is differentiation -- the ability to recognize what makes two concepts distinct.\footnote{We provide a more detailed discussion in Appendices \ref{sec:whatIsUnderstanding} and \ref{ap2}.} Accordingly, we evaluate in this paper models' capacity to discriminate between complex technological concepts that are linguistically similar yet functionally distinct. In our task, a model is presented with two texts (patent descriptions) and asked whether they describe the same underlying idea or two distinct ones. These texts are carefully selected to be semantically similar in language and structure, making the task deceptively difficult and surface-level heuristics ineffective. To succeed, the model must detect subtle differences in aspects such as purpose and mechanism. We choose patents as our primary testbed because they present a uniquely demanding environment; they are written in dense, legal-technical language, often strategically obfuscated, and their distinctions are adversarially verified by human patent examiners. This task requires fine-grained conceptual discrimination that goes well beyond the pattern matching (in strategically confusing texts), imitation, or memorization that often bypasses existing benchmarks \citep{mcintosh2024inadequacies, wu2024reasoning, davis2023benchmarks}.

Our results show that while frontier models display some competence, weaker models consistently miss key technological nuances. Building on this, we quantify whether failures on knowledge-intensive understanding tasks stem from missing knowledge or from having “lay-in” knowledge that models cannot reliably use. 

To measure the gap between working knowledge and lay-in knowledge, we prompt models to generate clarification questions to improve understanding about patents and compare baseline performance to two conditions: (i) questions and answers generated by the model itself, and (ii) questions and answers generated using external scientific information. We find that understanding failures are driven primarily by poor retrieval of lay-in knowledge: lay-in knowledge is substantially broader than working knowledge and, for our task, is nearly complete, containing roughly as much useful information as externally retrieved knowledge. Limited understanding is mainly a failure to access relevant knowledge at the right moment, not a failure to store it.

We then examine whether models differ in their ability to construct task-specific “databases” of usable cues. Smaller models tend to generate simpler, more broadly transferable cues that aid retrieval and downstream use, whereas larger models generate richer but more brittle and less generalizable cues. Together, these results point to a complementary strategy across scales: small models propose the scaffold, and large models do the heavy reasoning.

\paragraph{Contributions} (i) We introduce a scalable task and dataset to probe LLMs’ ability to deploy domain knowledge by operationalizing a core cognitive criterion of understanding - differentiation. We introduce a novel large-scale dataset of hard-to-separate USPTO patents (1.3M+ information/computer and 177K+ biomedical patents granted post-2015), where surface similarity is systematically misleading and success requires technical judgment. (ii) We introduce a diagnostic framework that separates missing from unused, "lay-in", knowledge. Applying this framework reveals a deployment bottleneck: models often have relevant knowledge but fail to activate it.

\section{Hard-to-Distinguish Patents}
\label{patent}
\begin{figure}[htbp]
    \centering
    \begin{center}
    \begin{subfigure}{0.24\textwidth}
        \centering
        \includegraphics[width=\linewidth]{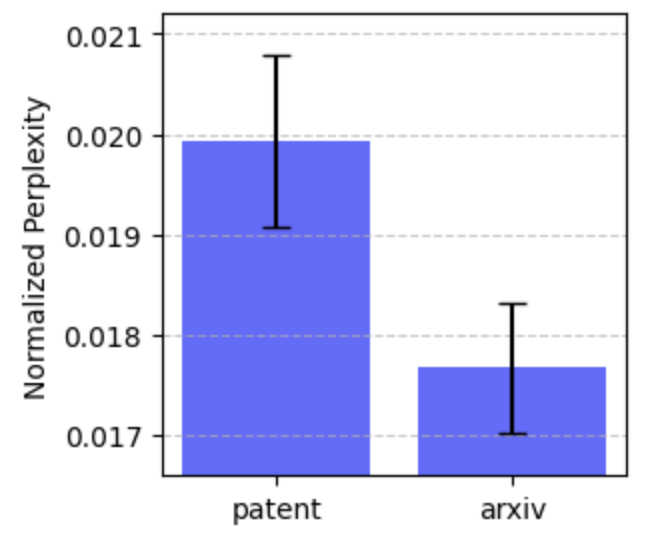}
        \caption{Patents are hard to read and understand}
        \label{fig:arxivVsPatents}
    \end{subfigure}
    \begin{subfigure}{0.23\textwidth}
        \centering
        \includegraphics[width=\linewidth]{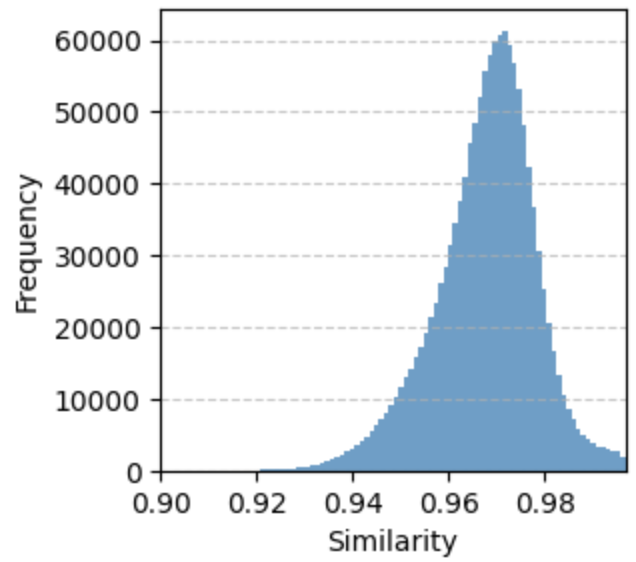}
        \caption{Patents and counterparts are hard to distinguish}
        \label{fig:patentsSimilarity}
    \end{subfigure}
    \end{center}
    \caption{Panel (a): Model perplexity of LLaMA-3.1-8B-Instruct on paper texts and patents descriptions. The texts of patents are much more confusing to LLMs compared to scientific papers. Panel (b): Histogram of the distribution of each patent's maximum cosine similarity to another (distinct) patent.}
    \label{fig:combined}
\end{figure}

The dataset consists of 1,319,184 information and computer technology and 177,903 biomedical patents. The designed task is less likely to be impacted by pretraining as shown in Appendix \ref{pretrain}. Several features make technology patents particularly well-suited for our test:

\textbf{First}, patents describe components of larger systems and thus occur in \textbf{dense clusters of near-duplicates} that are highly similar in structure and language but differ in a small number of substantively meaningful dimensions (e.g., constraints, use cases, or technical improvements) \citep{ouellette2012disclose}. These fine-grained distinctions - rigorously evaluated by human judges (see the third point) - directly test whether LLMs can move beyond surface similarity to recognize functional differences. We illustrate this pattern in Figure \ref{fig:patentsSimilarity}, which shows the distribution of similarities between one patent and its most similar -- but non-identical -- counterpart based on the textual embedding from \citep{ghosh2024paecter}, pretrained on patent-citation relations. The distribution is heavily concentrated at high similarity scores, suggesting that most patents have at least one semantically similar counterpart. We collect these similar but different patents and test whether LLMs can tell the nuances. In the Appendix \ref{ap3}, we also explore another type of "merging" task: whether a model can recognize a rewritten description referring to the same invention. We find that the differentiation task requires genuine conceptual linking and functional comparison, making it a strong testbed for evaluating LLM understanding in the technology domain. 

\textbf{Second}, patent documents are often \textbf{strategically obfuscated}, disclosing only what is legally required while minimizing substantive revelation. Consequently, their language is deliberately dense, formal, and legalistic. Readability studies show that patent texts demand substantially higher comprehension than comparably technical scientific prose due to jargon-heavy and legally constrained drafting conventions \citep{kong2023linguistic,ouellette2012disclose}. This linguistic complexity is further illustrated in Figure \ref{fig:arxivVsPatents}, which compares the model perplexity of LLaMA-3.1-8B-Instruct on 4,000 randomly selected computer science paper texts from ArXiv and 4,000 matched patent descriptions from the same years. Perplexity, while not a direct measure of understanding, reflects the model’s inability to predict the next word in a sequence. Consistently higher perplexity on patents suggests that they contain highly technical and domain-specific language that poses a greater challenge for LLMs to parse and comprehend. An example of paired patents and a related paper can be seen in Appendix \ref{ap5}.


\textbf{Third}, and most importantly, despite surface similarity, granted patents are known to be conceptually distinct because each has undergone \textbf{extensive expert review} by USPTO examiners. This process evaluates novelty, non-obviousness, and utility through detailed prior-art searches and comparative analysis. As a result, institutional validation provides a strong, domain-informed, human-expert-evaluated guarantee that different granted patents represent non-trivial conceptual distinctions, even when they appear nearly identical.


\section{Evaluating LLMs’ Understanding of the Technology Domain}

Hard-to-distinguish patent pairs (two patents show high text similarity but with different granted IDs) are presented to the LLM, which is asked whether the two texts describe the same invention.\footnote{See Appendix \ref{ap6} for full prompts.} For each prompt, the model returns two outputs: a binary label indicating “yes” (1) or “no” (0), and a confidence score on a scale from 0 to 10. Note that the expert-evaluated ground truth is always "they are different patents".

To construct our main measure of understanding, we aggregate LLM's output using a confidence-weighted voting scheme applied over three independent generations of the same prompt. If the total confidence associated with label “1” exceeds that of label “0,” the final label is set to 1; otherwise, it is 0. This aggregation method has been shown to represent the model's actual certainty, accelerate convergence toward the model’s stable judgment, and improve self-consistency relative to simple majority voting \citep{geng2023survey, taubenfeld2025confidence}. In addition to providing more reliable final decisions, confidence scores offer a fine-grained signal that helps distinguish between borderline and clear-cut cases. This is particularly important in our setting, where many patent pairs are deceptively similar—confidence levels can help identify when the model is genuinely uncertain \citep{geng2023survey}. The result of confidence-weighted voting correlates with majority voting without any confidence signals, and the correlation is 0.95. High confidence also well correlates with high judge accuracy, with a correlation of 0.92. Results using the majority voting can be found in Appendix \ref{ap8}. We also track changes in confidence. Throughout the experiments we use 5000 randomly sampled pairs to approximate the result on the full dataset as they are statistically sufficient. We find results on 1000, 5000, and 10000 randomly sampled pairs show a high correlation of over 0.98.

\begin{figure}[htbp]
    \centering
    \includegraphics[width=0.5\textwidth]{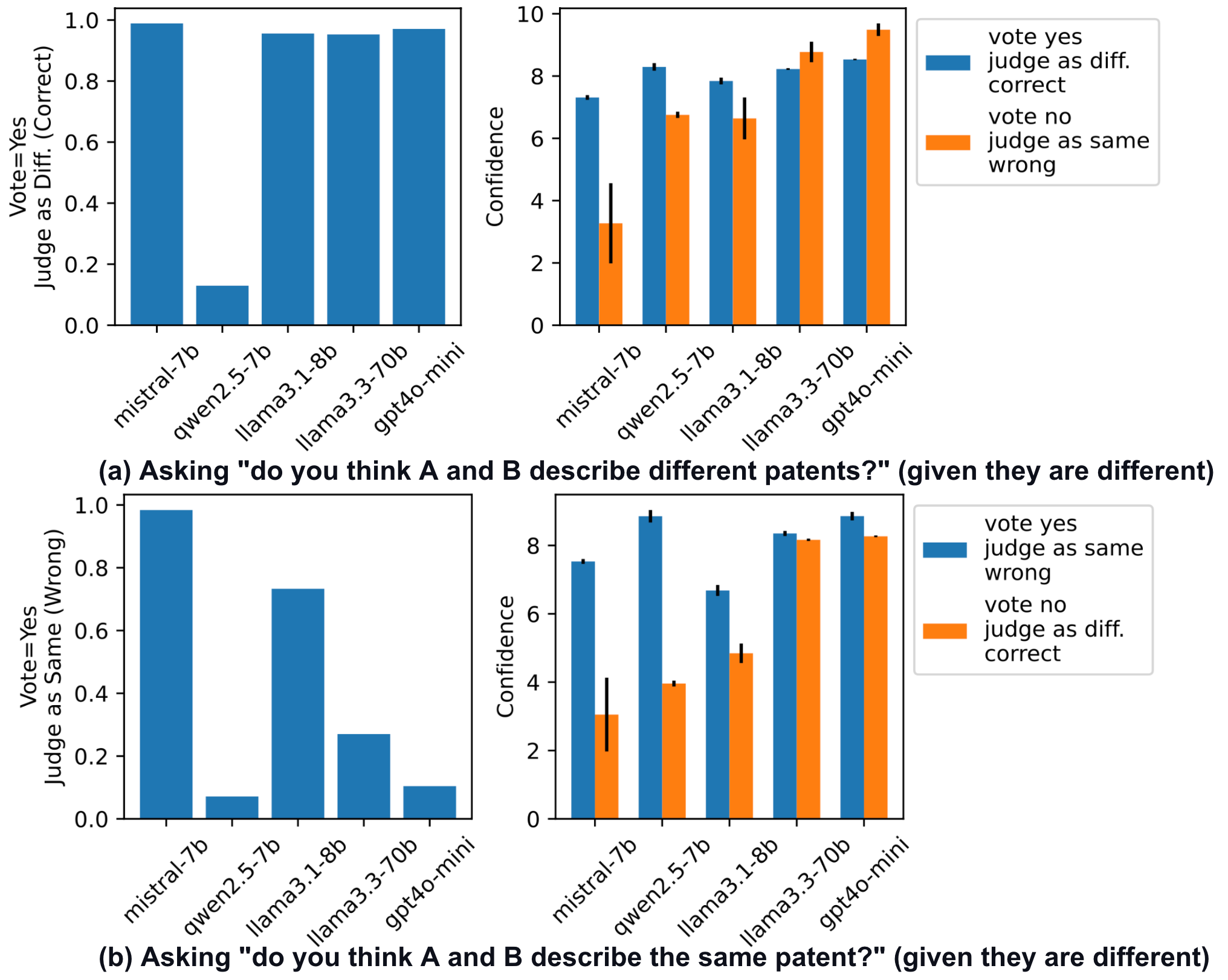}
    \footnotesize
    \caption{Blue bars always represent the answer "yes", orange bars always represent "no". Top row: Ask LLMs "Do you think they are different patents?", given that they are actually different. Blue: "yes" (they are different). Orange: "no" (they are the same). Bottom row: "Do you think they are the same patent?", given that they are actually different. Blue: "yes" (they are the same). Orange: "no" (they are different). Llama models are instruction-tuning versions.}
    \label{fig:llmUnderstanding}
\end{figure}

Looking at the top row of Figure \ref{fig:llmUnderstanding}, we evaluate LLM understanding by presenting two close-but-distinct patents and first asking <\textbf{whether they are different?}> Blue bars here always represent the answer "yes", and in this case, the proportion of responses "yes, they are different" (i.e., correct answer), along with their associated confidence, while the orange bars always denote the answer "no", and in this case, the responses "no, they are not different" (i.e., they are the same, incorrect answer). These results suggest that most models - with the exception of Qwen 2.5 7B - perform remarkably well; Mistral 7B, Llama 3.1 8B, and larger models like Llama 3.3 70B and GPT-4o mini all appear to identify the patents as distinct at a high rate (left panel, top row).

However, further investigation reveals that this seemingly high accuracy is superficial rather than a reflection of true understanding. We introduced a logically consistent inverse question in the bottom row of Figure \ref{fig:llmUnderstanding}: <\textbf{Are these patents the same?}> For a model to demonstrate a robust grasp of the nuances between these hard-to-distinguish patents, it should answer "yes" to the first prompt (are they different?) and "no" to the second (are they the same?). Interestingly, our results reveal significant response biases that undermine the top-row results. For instance, the Mistral model tends to answer "yes" regardless of the prompt's framing, whereas Qwen 2.5 7B shows a persistent tendency to answer "no". Furthermore, we observe a paradoxical "overconfidence" effect, where small models (Qwen 2.5 7B, Mistral 7B, and Llama 3.1 8B) always exhibit higher confidence when they answer "yes", and for larger models like Llama 3.3 70b and GPT-4o-mini, wrong answers show a higher confidence than correct ones.

We also find GPT-4o mini demonstrates the strongest capability, followed by Llama 3.3 70B and then Llama 3.1 8B. This is evidenced by the fact that while all three models frequently identify that the patents are different ("yes" responses
when asked if the patents are different) in the top row, GPT-4o mini maintains the lowest—and therefore most accurate—rate of "yes" responses when asked if the patents are the same in the bottom row. This suggests that as model scale and sophistication increase, so does the accuracy and the ability to maintain logical consistency when distinguishing between intricate technical documents.






\section{Experimental setups}


For failures in knowledge-intensive tasks, the most intuitive solution is to supplement the model's knowledge, with the most direct method being Retrieval-Augmented Generation (RAG). Based on this premise, we designed a set of control experiments to analyze why models fail. Our motivation is straightforward: we allow the model to self-diagnose by asking, "What specific background knowledge would help you better distinguish between these two patents?" \footnote{See Appendix \ref{ap6} for full prompts.}. We then tested two distinct conditions: (1) \textbf{Scientific RAG}: given the set of generated questions, supplying external scientific knowledge to address \emph{missing information}, if any. (2) \textbf{Self-QA}: given the same set of generated questions, LLMs answer their own questions, testing whether the model can bridge the gap using its own internal knowledge, indicating a \emph{utilization failure}. We then compare the content of these two types of answers and their impact on improving the judgment to determine the major drivers of failures. 

Specifically, we design two types of self-generated questions. The first, surface-level questions, target conceptual recall (e.g., What is X? What does Y do?), while the second encourages deeper understanding through questions about the integration of concepts, improvement, and the intent underlying invention. When the model is exposed to a pair of closely related patents, $P_1$ and $P_2$, it generates six questions $Q_{ij}$ for each, where $i \in {1,2}$ indexes the patent and $j \in {1,\dots,6}$ indexes the question. Three of these are surface-level questions, and three target deeper comprehension. Randomizing the order of questions does not impact the result much. The prompt for generating questions could be found in Appendix \ref{ap6}. 

In the Scientific-RAG setting, these questions guide the retrieval of relevant scientific knowledge. For external knowledge, we use parsed, pre-processed full texts from arXiv computer science papers. Using Google Scholar, we identify the top 10 papers relevant to the patents, segment each into overlapping 2500-character chunks (with 200-character overlaps), and treat these as candidate knowledge units. We embed both the questions and text chunks using the SPECTER2 model \citep{singh2022scirepeval} specifically for scientific documents pretrained on citation relations, compute cosine similarity, and select the top three relevant chunks per question. These are then passed to the LLM as external knowledge $Knowl_{ij}$. The model answers its own questions using the retrieved content $Knowl_{ij}$, producing scientific-augmented answers, and we feed the six scientific QA pairs for each patent to LLMs, plus the paired patent information, to let them judge again whether the two patents are the same. 

In the Self-QA setting, all experimental conditions remain the same as above, except that no external scientific retrieval is used. Instead, the LLM generates its own answers and then conditions its judgment on these self-generated QA pairs together with the patent information. 

\section{Results}\label{sec:results}

In this section we decompose the error rate into its components - how much of the error is due to missing information and how much is due to the inability to deploy self-knowledge. Section \ref{result2} shows our main result. In the following section, we analyze how model size affects question quality and judgment accuracy (\ref{result4}). Across the experiments below, we ask “do you think they are the same patent” as shown in Figure \ref{fig:llmUnderstanding}. All statistical significance analyses use the t-test. Ablation studies on the generated questions and retrieval quality can be found in Appendix \ref{ap:ablation}.

\subsection{Missing vs. Unused Knowledge}
\label{result2}

\begin{figure}[htbp]
    \centering
    \begin{subfigure}[b]{0.24\textwidth}
        \centering
        \includegraphics[width=\textwidth]{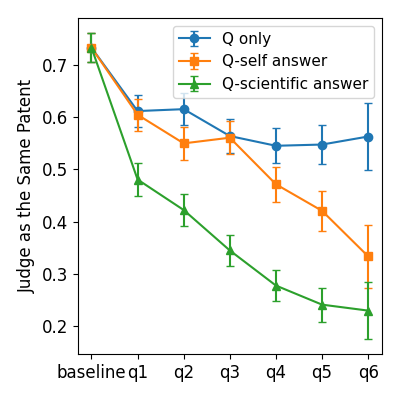}
        \caption{Self-talk helps improve judgment, but not as much as scientific QA.}
        \label{fig:know_know_llama8b}
    \end{subfigure}
    \hfill
    \begin{subfigure}[b]{0.23\textwidth}
        \centering
        \includegraphics[width=\textwidth]{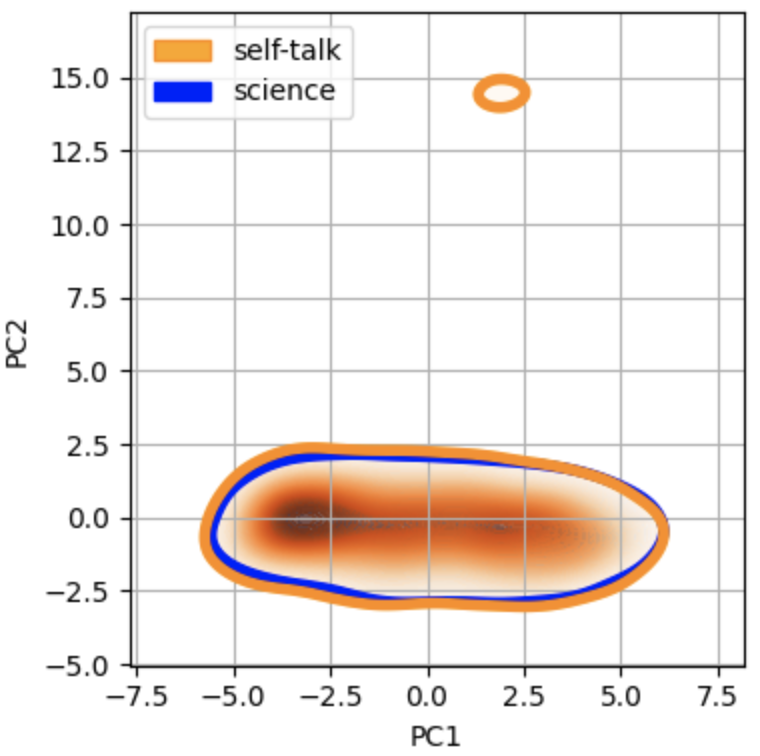}
        \caption{Scientific retrieval does not introduce more “content-level” new information.}
        \label{fig:subfig112}
    \end{subfigure}
    \caption{Panel a: The contribution to understanding decreases in the following order: scientific QA, self-QA, and question-only. $p$-values: $p_{q_1}$ to $p_{q_5}$ = 0.000, $p_{q_6}$ = 0.013 (self vs. scientific answers). Panel b: no significant difference could be found between scientific information and self-produced answers.}
    \label{fig:combined_majority111}
\end{figure}

Figure \ref{fig:know_know_llama8b} presents our results from the random sample of 5000 patent pairs, examining how questions and answers affect classification performance. The baseline represents the initial accuracy of Llama 3.1 8B without any self-questioning or scientific retrieval. At this baseline level, Llama 8B incorrectly classifies approximately 70\% of patent pairs as identical. Increasing the number of question–answer pairs leads to a more robust understanding.

The blue line demonstrates the impact of adding questions alone. This modest improvement suggests that simply posing understanding questions surprisingly helps the model perform better by providing structures and cues that aid text parsing and internal retrieval, though the effect of merely enumerating questions remains limited. The orange line reveals that adding self-generated answers significantly improves performance. This substantial gain implies that the model already contains useful knowledge for the differentiation task, but this knowledge is not being effectively utilized unless explicitly prompted. The green line shows the improvement achieved by providing scientifically retrieved answers. Importantly, error reduction is achieved more substantively through baseline to self-generated answers than self-generated answers to scientific ones, suggesting that the model contains the most useful information within its internal knowledge representation. However, the model lacks the ability to effectively access and apply this knowledge when tackling tasks that require such domain expertise.

We next explore the factors driving this performance difference between self-produced answers vs scientific answers, considering two possible explanations. First, LLMs function as compressed representations of knowledge but do not preserve the richness of information. Scientific and technological information tends to be sparsely and unevenly distributed in training corpora, and the compression process significantly degrades the richness of such underrepresented knowledge \citep{jiang2020skcompress, jaiswal2023compressing, hoang2023compressed, sun-etal-2024-head}. This implies that the \emph{“missing information” hypothesis largely does not hold}. The LLMs appear to internally store the relevant information; however, they either fail to effectively deploy it or suffer from a loss of informational richness due to compression. We thus also observe that cues only (questions) still help the deployment and dynamic use process within the compressed knowledge store. Second, LLMs may acquire new knowledge during external retrieval - meaning the external knowledge we provide is entirely unknown to the LLM and must be learned and incorporated during judgment. That implies that \emph{the “missing information” is the key.} 

We argue that the first explanation is likely the primary cause of the improvement we observe. The models have likely already encountered the scientific data we use as external sources: LLaMA 3.2 was released around October 2024\footnote{\url{https://huggingface.co/meta-llama/Llama-3.2-3B-Instruct}}, when approximately 86\% of the retrieved papers that generated scientific answers had already been published. 

To further investigate what type of information is contained in inference-time retrieval versus the model's internal knowledge, we analyze the answers by embedding all responses from both scientific sources and the LLM's self-generated answers using the model from \citep{ghosh2024paecter}, specifically pretrained on patent citations. To visualize and compare the distribution of embedding spaces, we reduce dimensionality to 2D and estimate boundaries using 5\% Gaussian kernel density estimation—drawing contours that enclose the densest 95\% of data points in each embedding set. If LLMs substantially incorporate new scientific content for answers by external retrieval, we would expect their embedding regions to diverge meaningfully from those derived from self-generated responses. However, as shown in Figure \ref{fig:combined_majority111} panel b, the embedding regions largely overlap. Notably, self-generated responses display a distinct outlier region, which we suspect reflects hallucinated content introduced by the LLM. We also qualitatively interpret these questions and answers as shown in Appendix \ref{qual}.

Despite occupying similar regions, these two answer types differ in several meaningful ways. Scientific answers average $62.1\pm1.8$ words and are presumably richer in detail and informativeness. In contrast, self-generated answers are notably shorter, averaging only $38.2\pm0.6$ words—approximately half the length of scientific responses. We therefore, interpret the results as the following insights.

\begin{tcolorbox}[colback=pink!25,colframe=black!75,title=The issue of compression]
Although in our settings LLMs largely possess the knowledge required for knowledge-intensive tasks, this knowledge is highly compressed, less detailed, and substantially less rich than the original scientific data from which it was learned.
\end{tcolorbox}

\begin{tcolorbox}[colback=pink!25,colframe=black!75,title=The issue of deployment]
Unless explicitly activated through mechanisms such as self-generated answers or structured elicitation cues (e.g., questions alone), the compressed knowledge in LLMs remains difficult to access and largely unavailable for dynamic deployment.
\end{tcolorbox}

This perspective also challenges an assumption of RAG: that supplying retrieved documents can remedy knowledge deficiencies at inference time. If performance gaps arise from limitations in knowledge deployment or integration - rather than simple absence - retrieval may be insufficient. Retrieved content can remain shallowly integrated or misaligned with internal representations. Thus, RAG treats knowledge access as the bottleneck \citep{zhou2025retrieval, cheng2025survey}, whereas our findings suggest that knowledge representation and dynamic use are equally, if not more, critical. Improving performance may therefore require advances in how models internalize retrieved information, not merely expanding input.



\subsection{It's Not Only What to Ask, but Who Asks It}
\label{result4}

\begin{figure*}[h]
    \centering
    \includegraphics[width=\textwidth,height=0.7\textheight,keepaspectratio]{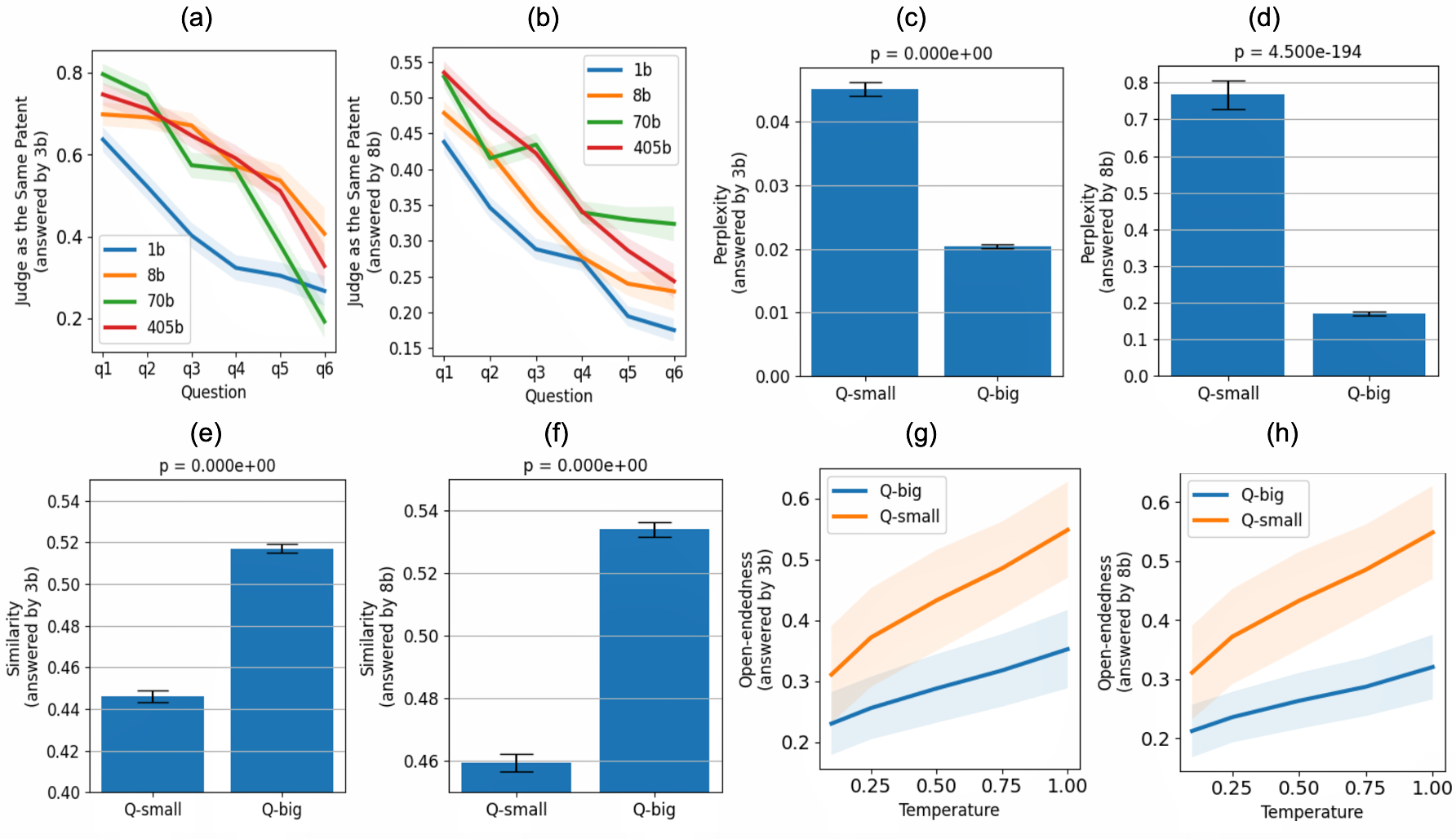}
    \caption{Question models: llama 1b, 8b, 70b, 405b. Answer models: 3b and 8b. Panels a and b: Self-generated questions are compatible with the model’s own answering capacity. Questions generated by smaller models are compatible with larger answer models but not the converse. "Compatible" means a lower misjudgment rate (judging as "same patents"). Panels c to h: mechanisms. Panels c and d: higher perplexity when answering small models' questions. Panels e and f: lower similarity between [small models' questions + answer models' answers] and [the patent texts], implying searching for more new information when answering small models' questions. Panels g and h: smaller models' questions are more open-ended (the SD of embedding distances of each answer from the mean of all answers).}
    \label{fig:smallerModelsFigure}
\end{figure*}

A transferring question emerges: are the most useful questions those that \emph{the same} model would ask itself, or can questions authored by a different model—smaller or larger—better align the answerer’s knowledge with the task? We next test how question–answer sets travel across scale and where misalignment helps or hurts.

The answer depends on which model is doing the answering. We test two models—LLaMA 3B and 8B—on questions generated by four others: LLaMA 1B, 8B, 70B, and 405B. We select 3B and 8B as answerer models because the 1B model produces low-confidence responses (94.4\% with a confidence score of 0), indicating little understanding of content. In contrast, the 70B and 405B models already achieve strong performance without an external scientific context, leaving little room for improvement. The 3B and 8B models provide a more informative middle ground: they show some understanding but still benefit from superior prompts.

As shown in Figure \ref{fig:smallerModelsFigure} panels a and b, we find that the 3B model performs best when answering questions from the 1B model, and the 8B model benefits from questions posed by both the 1B and 8B (i.e., misjudge rate (judging as "same patents") is lower than questions from bigger models). This asymmetry reveals a critical insight: questions from larger models do not consistently help smaller models, and often introduce confusion. A more knowledgeable model cannot fully substitute for a smaller model’s understanding, suggesting that knowledge extraction is not smoothly transferable across model scales \citep{li2024hindsight}. At the same time, we observe that questions from smaller models often align well with the answering capacity of larger models. This may be because larger models can readily interpret, build on, and give a better answer to smaller models' questions. Large model questions appear to go beyond the bounds (and dimensionality) of smaller models' understanding and answering capacity.

To probe this further, we generate 10,000 question-scientific answer pairs using the 3B and 8B models (the answer model), based on questions from either smaller or larger models relative to the answer models' own size (smaller or bigger question models), resulting in 40,000 total pairs. We then measure textual similarity between each QA pair and its associated patent text. When a larger model answers questions from a smaller model, the resulting QA content is consistently less similar to the original patent text, suggesting that small models' questions bring in more external scientific information to improve the quality, texture, and richness of knowledge beyond the raw patent texts (Figure \ref{fig:smallerModelsFigure} panels e and f). Small model questions ``discover more'' \textbf{external} information from the sources they search. We also re-feed the scientific QA pairs into the 3B and 8B models and record their model perplexity. This lower similarity predictably activates more \textbf{latent} reasoning in the answering models, as reflected in their answer's higher perplexity (Figure \ref{fig:smallerModelsFigure} panels c and d). This may stem from how smaller models phrase their questions. With limited capacity, their questions tend to have two important features: (1) Small model questions are more fundamental — 68\% of 1B’s questions contain “what is” or “what are” (usage/purpose of components, concepts, and so on), compared to only 44\% for 405B. They are precisely the fundamentals of judging patents. (2) Smaller models tend to ask shorter, less contrastive, and more open-ended questions, potentially yielding higher information entropy (Figure \ref{fig:smallerModelsFigure} panels g and h). A 1B model's questions average 16.7 words, compared to 23.0 for a 405B model. We further sampled 200 questions from both smaller and larger question models. The answer model (3B or 8B) was prompted to respond to each question ten times across temperatures from 0.1 to 1, controlling for the answer model’s underlying determinism. We quantify “open-endedness” as the embedding distance of each answer from the mean of all answers (standard deviation). Questions from smaller question models elicit more varied answers.

\section{Final Remarks}

We reframe LLM failures not as gaps in knowledge acquisition, but as limitations in knowledge deployment. Using a patent differentiation task that requires the dynamic use of knowledge, we show that models often possess relevant knowledge yet fail to activate it, whereas external retrieval yields diminishing returns when utilization—not access—is the primary bottleneck. Complementarities across model scales further suggest hybrid strategies that combine structural scaffolding from smaller models with deeper reasoning from larger ones. This work brings insights into the right way of building science-powered LLMs for automating the assessment of innovation.

\section*{Limitations}
\label{ap:limitation}
Although our current setup uses scientific retrieval from arXiv papers, a more promising direction is to evolve the retrieval process itself. Rather than relying on fixed sources or searching questions from scientific databases, LLMs could learn to refine or even generate better retrieval queries via reinforcement learning, active learning, or model-in-the-loop optimization based on science or what they have learned from prior reasoning steps. An agentic framing - where models, like a baby sensing the world, autonomously refine their questions and knowledge, decide whether to retrieve information, evaluate whether to accept or reject what is retrieved, and determine how much knowledge to seek for one question/how many questions to ask - could enable deeper forms of exploration beyond single-pass self-questioning.

The exact nature of ``resolution'' in distinguishing patents remains underdefined. That is, even when models succeed at effectively splitting, it is not always clear what semantic threshold they are using. Are they resolving based on functionality, application, architecture, or terminology? Future work could explicitly model and evaluate these questions to understand what kinds of conceptual distinctions LLMs are truly capable of making by probing their internal, layer-wise conceptual representations and dynamics. It will also be interesting to look into differences in judgment performance across patent classes and categories, as LLM internal knowledge likely varies across fields.

\section*{Ethics Statement}
This work complies with the Code of Ethics. We used AI assistants to expedite the coding process. All AI-generated code snippets were carefully reviewed and verified by the first author prior to incorporation. For manuscript preparation, AI assistants were used exclusively for grammar checking. No external ethics review was required, as the experiments involved only the analysis of patent texts and therefore posed no risk to individuals across any demographic or geographic groups.
The patent data are licensed under the Attribution 4.0 International license (see \url{https://patentsview.org/rules-of-conduct?utm_source=chatgpt.com}), which permits copying and redistribution in any medium or format for any purpose, including commercial use. The data are publicly available through the USPTO; our contribution was limited to collection and cleaning. All experiments were conducted using two NVIDIA A100 GPUs with 40 GB of memory each.

\newpage

\bibliography{references}

\newpage

\appendix

{\Huge Appendix}

\section{Understanding Through Differentiation}\label{sec:whatIsUnderstanding}

Understanding is an ability that goes beyond merely recalling isolated facts or performing procedures. It fundamentally involves grasping the relationships that connect concepts and principles within a domain. This requires both an appreciation of and the ability to uncover underlying structures, such as causal links, correlations, hierarchies, or analogical parallels, which distinguish true comprehension from sophisticated pattern matching or associative recall. Evaluating an LLM's understanding, therefore, requires methods that probe its capacity to navigate these relational structures, moving beyond tests of surface-level knowledge.

Cognitive science, from \citep{gibson1969perceptual} through \citep{tversky1977features} and Gentner’s structure-mapping work \citep{markman2000structure} to recent relational-reasoning research \citep{alexander2018relational}, demonstrated that the ability to perceive, represent, and reason about crucial semantic relationships is built upon two complementary foundational processes: differentiation, contrasting, or ``splitting'', and generalization, merging, or ``lumping''. Differentiation is the capacity to distinguish concepts, entities, or situations, recognizing the features or relations that make them distinct -- essentially, identifying what makes 'A' different from 'B'. Without this ability to draw meaningful distinctions, even simple relationships cannot be properly conceptualized. Generalization, conversely, involves identifying commonalities across different instances, abstracting shared properties or underlying structures, and treating distinct items as equivalent for certain purposes, such as belonging to the same category or representing the same concept. These processes are fundamental not just for forming basic concepts, but for constructing the complex relational knowledge that constitutes deep understanding in any domain. 

Moreover, the ability to differentiate and generalize remains vital for constructing and refining our understanding of relationships between concepts, especially causal ones, which serve as the backbone of understanding. To explain why something occurs, we must first detect patterns: when does outcome $B$ follow event $A$, and when does it not? This requires the capacity to distinguish $A$ from alternatives like $A'$, and to notice that $B$ follows $A$ but not $A'$. Such contrasts form the basis for counterfactual reasoning, allowing us to imagine what would happen if everything about $A$ and $A'$ were held constant except for one defining feature—if changing that feature flips the outcome, it signals a likely cause. At the same time, we must also recognize when observations are similar in respects that matter. Only by treating $A$ and $A'$ as alike for relevant purposes, and then seeing whether their outcomes differ, can we meaningfully generalize and say that “$A$ causes $B$.” This dual capacity—to detect when things are different and when they are effectively the same—is what underlies both the identification and the generalization of causal claims. Contemporary computational models and econometric frameworks reflect this structure explicitly, emphasizing the need for clear conceptual boundaries and controlled variation \citep{pearl2009causality, lake2017building, heckman2022causality}. Developmental research similarly shows that even young children rely on these kinds of structured comparisons to infer causes \citep{gopnik2012causal}. 

LLMs have demonstrated a remarkable ability to understand, interact with, and respond to human language \citep{wang2018glue}, \citep{wang2019superglue}, \citep{openai2023gpt4}, \citep{brown2020language}. Nevertheless, understanding language or recalling facts alone does not guarantee understanding higher-order concepts or the construction of a world model \citep{vafa2024evaluating} that describes and explains how things relate to one another. The discussion above highlights that for LLMs to demonstrate comprehension of high-level concepts, they must be able to discern similarities and differences between complex concepts. Consequently, measuring an LLM's proficiency in discerning similarity and difference within highly specific concepts and fields offers a powerful proxy for evaluating its understanding of the conceptual foundations and landscape of these fields.

\section{Related Literature}\label{ap2}

\textbf{Relations to current benchmarks}: LLMs have achieved impressive scores on many NLP benchmarks, often giving the illusion of human-level reasoning or understanding. But researchers caution that excelling at a benchmark is not equivalent to possessing the general ability it is named after \citep{mcintosh2024inadequacies, wu2024reasoning, davis2023benchmarks}. High performance may reflect superficial pattern matching or memorization rather than true conceptual comprehension. In particular, current evaluation datasets frequently allow models to exploit statistical associations or training data overlaps to get the right answers without genuine understanding \footnote{https://www.sciencenews.org/article/ai-understanding-reasoning-skill-assess}. Even when test content is novel, models can exploit formatting quirks. Current benchmark agreement tests are strongly influenced by benchmark-orthogonal factors such as question format \citep{wu2025mapping}. Alzahrani et al. \citep{alzahrani2024benchmarks} showed that minor changes to the evaluation format – like shuffling multiple-choice answer order – can significantly alter MMLU performance and even flip model leaderboard rankings by several positions. Such brittleness indicates models are learning task-specific tricks (e.g., defaulting to a particular choice position or prompt wording) rather than demonstrating robust knowledge application. Minor wording tweaks could cause inconsistent outputs, suggesting that models are latching onto spurious lexical cues instead of grasping underlying semantics \citep{arakelyan2024semantic}. Aggregate metrics in existing benchmarks often obfuscate key information about where models tend to succeed or fail, broadcasting an inflated view of LLM capabilities \citep{burnell2023rethink}. In sum, many current benchmarks are static and shortcut-prone, failing to truly assess whether an AI understands concepts or is simply leveraging surface regularities. Recognizing these deficiencies, researchers have proposed more diagnostic evaluations to probe the conceptual depth of LLMs, such as contrastive evaluation using paraphrase \citep{asthana2024evaluating} or contrast and counterfactual sets \citep{lewis2024using} \footnote{Some have expressed concern that even discriminative capacity falls prey to the ``Generative AI Paradox'', wherein models can produce fluent, even expert-level outputs that far exceed their actual comprehension of the content \citep{west2023generative}.}.

\textbf{Organizing internal and external knowledge}: A core challenge in LLM understanding is how to activate the vast latent knowledge embedded in the model’s billions of parameters. Modern LLMs are trained on enormous corpora and ostensibly “know” a great deal, but this knowledge is stored in a highly compressed, distributed form not always accessible on demand \citep{borgeaud2022improving, hoang2024simple, hoang2023compressed, jaiswal2023compressing, pan2025understanding}. Indeed, there is a tension: an LLM may contain the facts needed for a task but still fail to recall them or apply them correctly when prompted.
The model might skillfully "reason" with the information it does recall, yet still hallucinate a critical missing fact or overlook a subtle piece of world knowledge. This limitation is evident in tasks requiring niche or up-to-date information: the needed knowledge might exist somewhere in the model’s parameters, but eliciting it reliably through prompts is non-trivial \citep{schulhoff2024prompt, gopnik2012causal}. To address these issues, researchers have proposed retrieval-augmented generation (RAG) techniques, which marry LLMs with external knowledge sources. Rather than relying solely on the model’s compressed memory, a retrieval module fetches relevant documents (e.g., from Wikipedia or scientific literature) to provide the model with explicit evidence \citep{prince2024opportunities}. Lewis et al. \citep{lewis2020retrieval} demonstrated the promise of this approach by generating more specific, diverse and factual language than a state-of-the-art parametric-only seq2seq baseline in NLP tasks. Borgeaud et al. \citep{borgeaud2022improving} introduced the RETRO (Retrieval-Enhanced Transformer) model. A RETRO model of only 7 billion parameters matched the performance of a 178B parameter model without retrieval. He et al. \citep{he2022rethinking} showed that an LLM with a retrieval step produces more faithful, grounded understanding – their “rethinking with retrieval” method led to answers that stayed consistent with retrieved evidence, improving transparency and correctness. In general, augmenting LLMs with retrieval has been found to boost factual accuracy, reduce hallucinations, and expand the range of answerable queries. LLMs compress information, and so lose rare information in the ``tail'' of the distribution, leading to phenomena like model collapse when trained on recursively generated data~\citep{shumailov2024ai}. By searching through uncompressed information with RAG, LLMs also access critical and potentially rare information for making detailed decisions about complex concept similarities and differences.

\textbf{Contrastive learning}: Our measures are also related to recent work on contrastive learning, which aims to improve model representations by explicitly training them to recognize semantic similarity and dissimilarity \citep{he2020moco, gao2021simcse, chen2020simclr, cohan2020specter}. Contrastive methods work by pulling representations of semantically related inputs closer together while pushing those of unrelated inputs apart, fostering a latent space that reflects nuanced conceptual structure. While our evaluation does not involve contrastive training \textit{per se}, it is motivated by a similar principle: if a model can consistently discern between distinct concepts, it demonstrates a deeper form of understanding—one likely to translate into stronger performance on downstream tasks.

\section{Judging Close Patents VS. Rewritten Patents}\label{ap3}

\begin{figure}[htbp]
     \begin{center}
     \begin{subfigure}{0.23\textwidth}
         \centering
         \includegraphics[width=\linewidth]{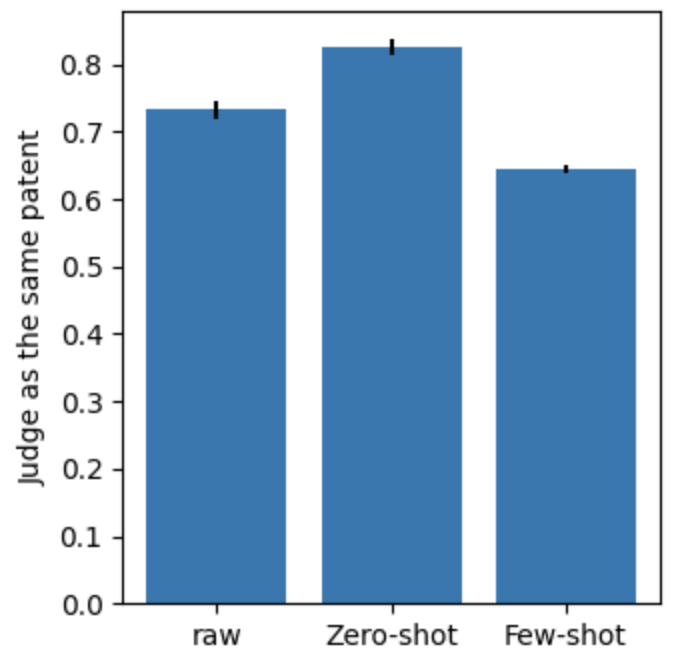}
        \caption{Few-shot CoT fails to help distinguish between patents with meaningful technical differences.}
        \label{fig:split_cot}
   \end{subfigure}
    \begin{subfigure}{0.23\textwidth}
        \centering
        \includegraphics[width=\linewidth]{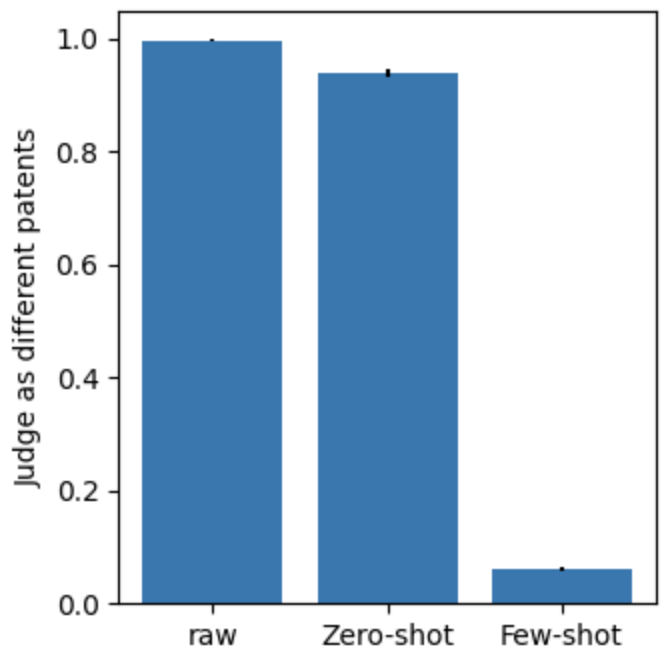}
        \caption{Few-shot CoT helps identify similarities across patent texts and their rewritten versions.}
        \label{fig:merge_cot}
    \end{subfigure}
    \end{center}
    
    \caption{Few-shot CoT reasoning can distinguish the rephrased patent texts (rewritten by the GPT-o3 model), as shown in the right panel (lower misjudge rate), but cannot distinguish their counterpart (another unique patent, as shown in the left panel). Y-axis always represents the response of "yes", i.e., the error rate (Left: yes when asking "are they the same" for close-but-different patent pairs. Right: yes when asking "are they different" for rephrased patents.)}
    \label{fig:combined}
\end{figure}

Following on prior sections that focus on distinguishing deceptively similar patents (the splitting task), in this section, we explore whether and how LLMs are able to merge identical but paraphrased patent ideas (the merging task). Our approach to measuring LLMs’ ability to generalize and identify the same ideas across distinct articulations is to generate rephrased versions of each patent text using the OpenAI ChatGPT o3 model with the prompt “rephrase the text, but keep its meaning.” We then task the LLaMA-3.1-8B-instruct model with identifying whether a patent text and its paraphrase describe different interventions (the correct answer: no). In raw (zero-shot) judgments, the model performed poorly on both merging and splitting tasks. We then introduced zero-shot Chain-of-Thought (CoT) prompting (“think step-by-step”) and few-shot CoT, where we supplied examples of both merging (paraphrased) and splitting (semantically similar but distinct) patent pairs. Few-shot CoT substantially improved the model’s accuracy in recognizing paraphrased texts as equivalent—but had minimal effect on its ability to distinguish between closely related yet different inventions.

This pattern suggests that more reasoning helps the model overcome basic semantic similarity checks and identify paraphrases, likely by encouraging step-by-step alignment at the surface level. Our hypothesis is that the model passes the rephrasing test primarily through pattern matching and semantic overlap, rather than through true conceptual integration. The merging test is relatively easy in the sense that our rewritten version keeps key technical terms unchanged but only changes how the sentence is structured, such that shallow strategies suffice. In contrast, differentiating between semantically similar but substantively different patents remains hard, because it requires more than aligned phrasing—it demands conceptual depth. 

In principle, a more advanced rephrasing model—one that intentionally avoids superficial alignment and instead rewrites concepts in deeper, functionally equivalent ways—might push the LLM closer to the more difficult task of distinguishing near matches. Designing such a rephrasing system is far from trivial. Thus, while our current rewrite approach does not fully obscure surface similarity, the model’s success only when the task is easy reinforces the motivation behind our task: LLMs rely heavily on surface-level semantic similarity, rather than demonstrating a deeper understanding of the underlying concepts.

\section{The Patent Dataset}\label{ap5}

The computer science patent dataset comprises 1,319,184 patent pairs in total, drawn from five Cooperative Patent Classification (CPC) classes: H05 – Electric techniques not otherwise provided for; H04 – Electric communication technique; G06 – Computing; Calculating; Counting; H03 – Basic electronic circuitry; and G11 – Information storage. Within these, the patents are further categorized into 41 CPC subclasses: ['H05B', 'H04L', 'H05H', 'H05K', 'G06V', 'G06N', 'G06T', 'H04N', 'G06F', 'G06Q', 'H03M', 'G06K', 'H04B', 'H04R', 'G11B', 'G11C', 'H04M', 'H03B', 'H03F', 'H03G', 'H03H', 'H03K', 'H03L', 'H04H', 'H04J', 'H04K', 'H04W', 'H04Q', 'H04S', 'H05F', 'H05G', 'H03C', 'H03J', 'G06E', 'H03D', 'G06G', 'G06M', 'G06C', 'H05C', 'G06J', 'G06D']. The 177,903 biomedical patents are drawn from USPTO classes C01, C07–C14, and C21–C30. Out of all matched patent pairs in our dataset, 82.55\% share the same CPC class, and 68.10\% share the same, more specific CPC subclass. In contrast, only 10.50\% of the pairs were granted in the same year.

This is an example patent pair where all LLaMA 3.1 instruct models — 8B, 70B, and 405B - fail to distinguish. We also show a scientific paper abstract to help readers perceive the distinct writing styles between patents and papers. Reference indexes are removed.

\textbf{Patent 1}: \textit{A wireless device is configured for use in a wireless communication system. The wireless device is configured to receive, from a radio network node, a multicast or broadcast transmission and downlink control information indicating a number of repetitions with which the multicast or broadcast transmission is transmitted. The wireless device is also configured to operate according to a negative acknowledgement only feedback scheme in which the wireless device is configured to transmit a negative acknowledgement to the radio network node if decoding of the multicast or broadcast transmission fails using the indicated number of repetitions and to refrain from transmitting a positive acknowledgement to the radio network node if decoding of the multicast or broadcast transmission succeeds.}

\textbf{Patent 2}: \textit{Methods, systems, and devices for wireless communications are described that support feedback for multicast communications. A user equipment (UE) in a group of UEs may receive multicast information from a base station in a downlink (DL) transmission scheduled by a DL grant. The DL grant may also indicate a set of uplink resources for the group of UEs to use to transmit feedback. The UE may attempt to decode a message containing the multicast information in the DL transmission. If the UE determines the message was successfully decoded, the UE may send no feedback to the base station. If the UE determines the message was not successfully decoded, the UE may transmit a feedback message to the base station to indicate the message was unsuccessfully decoded. The base station may monitor for the feedback message and determine to retransmit the multicast information if the feedback message is received.}

\textbf{An Arxiv paper including the answer to questions about judging the sampled patent pair}: \textit{The Third Generation Partnership Project (3GPP) has recently published its Release 16 that includes the first Vehicleto-Everything (V2X) standard based on the 5G New Radio (NR) air interface. 5G NR V2X introduces advanced functionalities on top of the 5G NR air interface to support connected and automated driving use cases with stringent requirements. This paper presents an in-depth tutorial of the 3GPP Release 16 5G NR V2X standard for V2X communications, with a particular focus on the sidelink, since it is the most significant part of 5G NR V2X. The main part of the paper is an in-depth treatment of the key aspects of 5G NR V2X: the physical layer, the resource allocation, the quality of service management, the enhancements introduced to the Uu interface and the mobility management for V2N (Vehicle to Network) communications, as well as the co-existence mechanisms between 5G NR V2X and LTE V2X. We also review the use cases, the system architecture, and describe the evaluation methodology and simulation assumptions for 5G NR V2X. Finally, we provide an outlook on possible 5G NR V2X enhancements, including those identified within Release 17.}

\section{Prompts}\label{ap6}
\begin{tcolorbox}[breakable, colback=gray!10!white, colframe=gray!80!black, title=Prompt for Question Generation]

\textbf{System Prompt:}
You are an expert in patent comprehension. Your task is to generate structured questions that assess a reader’s background knowledge necessary to understand a given patent. The questions should focus on foundational concepts, principles, and applications relevant to the patent’s domain without explicitly referencing the patent itself. Ensure the questions are framed generally and test for domain knowledge rather than details specific to the patent.

The questions should follow Bloom’s Taxonomy and be categorized into three levels:

1. Remembering: Questions that assess the reader’s ability to recall key technical concepts, definitions, and fundamental principles relevant to the patent’s domain. These should focus on identifying and defining key terms, recognizing fundamental components, and understanding core principles of related technologies.

2. Understanding: Questions that assess the reader’s ability to explain how different elements of similar technologies function and interact. These should encourage summarization, interpretation, and explanation of related technical concepts and their roles in broader systems.

3. Applying: Questions that evaluate the reader’s ability to apply their knowledge by solving problems, making predictions, or considering real-world applications of the broader technological principles underlying the invention. These questions should be scenario-based and encourage practical thinking.

Do not directly reference the patent text in any question. Instead, ensure all questions test knowledge that would be useful for understanding the patent without explicitly addressing its content. Maintain a structured and logical flow from basic recall to deeper conceptual understanding and practical application.

\textbf{Basic Level Question Prompt:}

You are given the following patent:

Patent Title: \{patent\_title\}

Patent Text: \{patent\_summary\}

Generate a set of \{qnum\} questions that test a reader’s ability to recall fundamental concepts, key terms, and basic components necessary to understand the given patent. The questions should focus on defining terms, identifying parts, and recognizing the primary function of the described technology.

Output Format (JSON):

Your response should be formatted as valid JSON:

\{\{

   ``1'': ``question1'',
   
   ``2'': ``question2'',
   
   ``3'': ``question3''
   
\}\}

\textbf{Conceptual Level Question Prompt:}

You are given the following patent:

Patent Title: \{patent\_title\}

Patent Text: \{patent\_summary\}

Generate a set of \{qnum\} questions that assess a reader’s comprehension of the given patent text by requiring them to explain how different elements of the invention work together. The questions should encourage the reader to summarize the design, describe interactions between components, and interpret the intended improvements of the invention.

Output Format (JSON):

Your response should be formatted as valid JSON:

\{\{

   ``1'': ``question1'',
   
   ``2'': ``question2'',
   
   ``3'': ``question3''
   
\}\}

\end{tcolorbox}

\begin{tcolorbox}[colback=gray!10!white, colframe=gray!80!black, title=Prompt for Answering Questions with Scientific Knowledge]

You are a scientific reasoning assistant.

Use the following context to answer the question. Focus only on the information provided.

Context: // retrieved scientific knowledge

\{context1\}

\{context2\}

\{context3\}

Question: 

\{question\}

1. Identify and summarize key points from the context relevant to the question.

2. Try your best to answer the question.

Return your answer in the following JSON format:

\{\{

``answer'': ``your answer'' //your answer to the question

\}\}

\end{tcolorbox}


\begin{tcolorbox}[colback=gray!10!white, colframe=gray!80!black, title=Prompt for the Splitting Task Judgment]

You are a patent judge. Below are two patent texts with extra information.

Your task: judge whether they describe the same patent.

Patent 1: 

\{patent1\}

\{qa\_block1\}

Patent 2: 

\{patent2\}

\{qa\_block2\}

Strictly return your concise answer in the following JSON format:

\{\{

  ``label'': 1 // 1 if they describe the same patent, 0 if not
  
  ``confidence'': 10 // your confidence score from 0 to 10, with 10 representing the highest confidence
  
\}\}

// Note:

// This applies to the major experiments in the paper: "Do you think they are the same patent?" — when, in fact, the patents are judged to be *different* by human experts.

// Invert this setup: Replace "same" with "different", and return 1 if the LLM concludes that the patents are *different*. This yields an alternative version of the experiment (See Section 3 Figure 2 top row in the paper).

\end{tcolorbox}
\begin{tcolorbox}[colback=gray!10!white, colframe=gray!80!black, title=Prompt for Self-talk Answering]

Try to answer the question.

Question: \{question\}

Return your concise answer in the following JSON format:

\{\{

"answer": "your answer" // your answer to the question

\}\}

\end{tcolorbox}

\clearpage

\section{Results for Majority Voting Schema}\label{ap8}


\begin{figure}[h]
    \centering
    \includegraphics[width=0.48\textwidth]{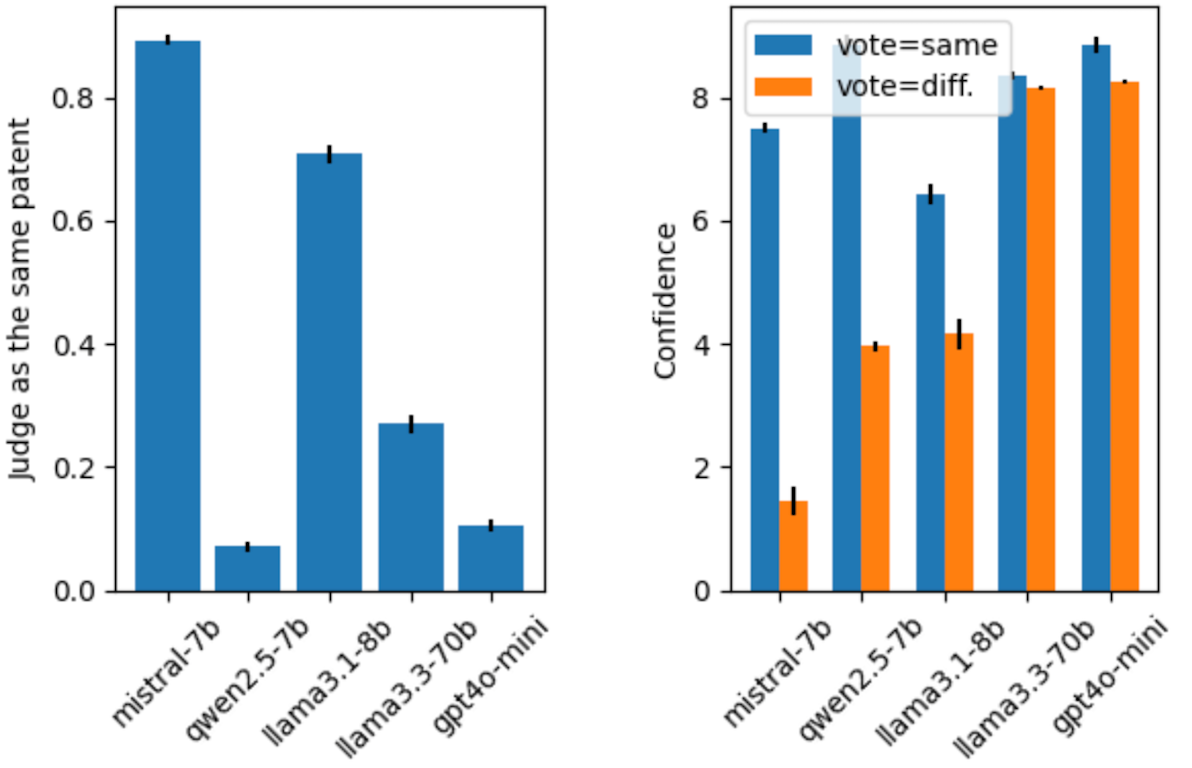} 
    \caption{Ask LLMs "Do you think they are the same patent?", given that they are actually different. The judge is using majority voting, and 1 = same (wrong answer). Llama models are instruction-tuning versions.}
    \label{fig:example1}
\end{figure}

\begin{figure}[h]
    \centering
    \includegraphics[width=0.48\textwidth]{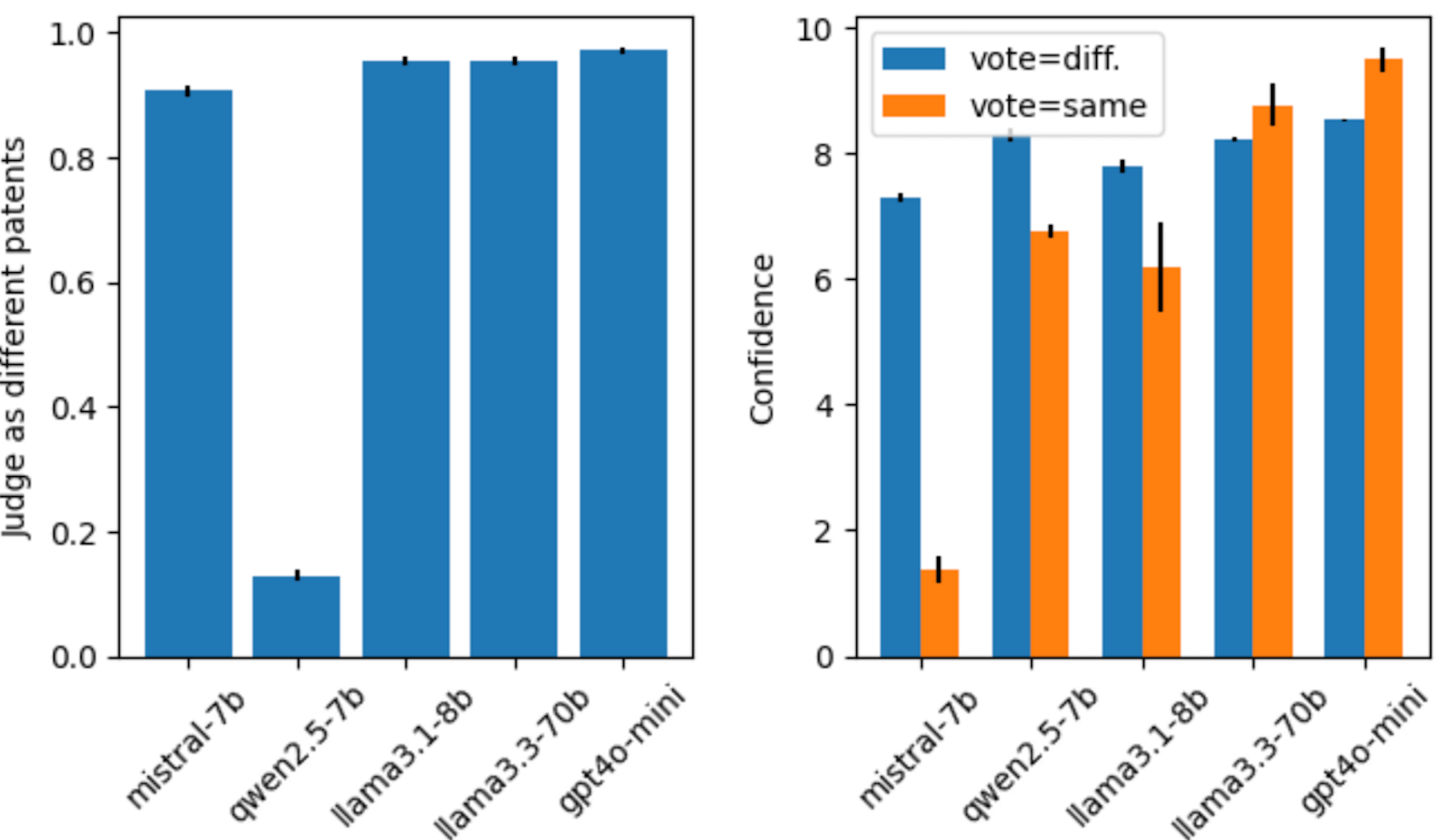} 
    \caption{Ask LLMs "Do you think they are different patents?", given that they are actually different. The judge is using majority voting, and 1 = yes (correct). Llama models are instruction-tuning versions.}
    \label{fig:example2}
\end{figure}




\begin{figure}[htbp]
    \centering
    \includegraphics[width=0.48\textwidth]{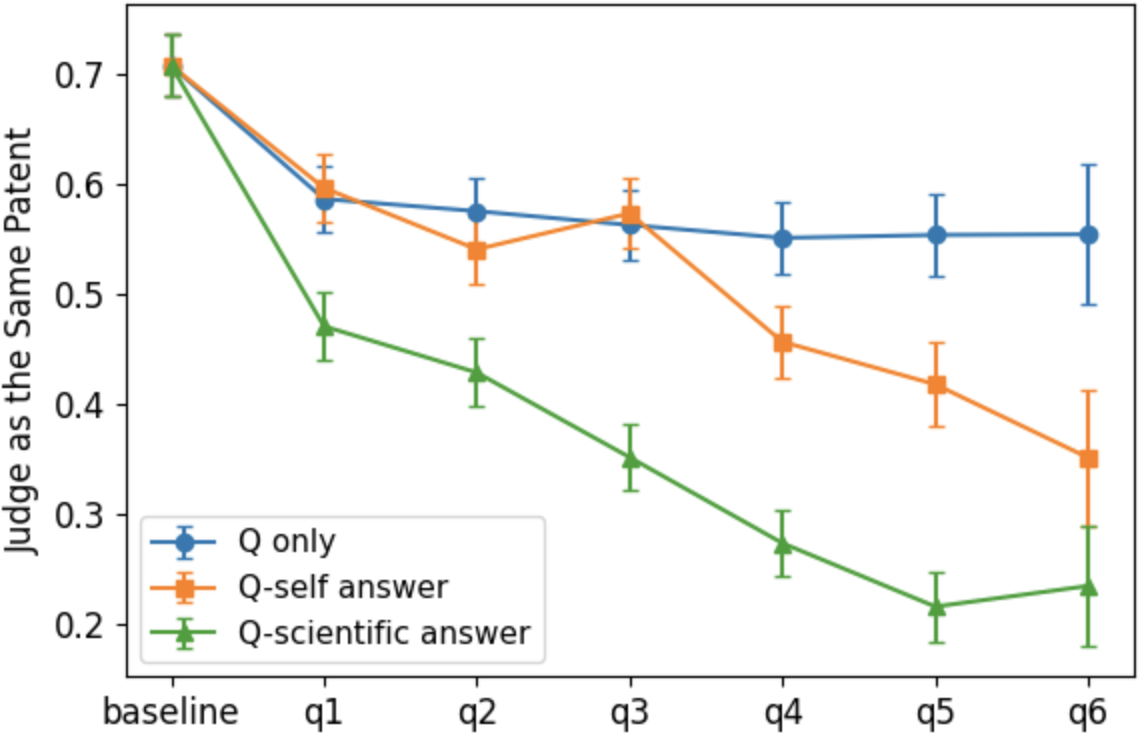}
    \caption{Asking Llama 3.1 8B instruct "Do you think they are the same patent?" on actually different patents (1 = wrong answer "yes"). LLM judges patents using \textbf{majority voting}. These results are statistically significant: Self-talk still helps improve the baseline judge but not as good as scientific answers: $q_1$ to $q_5$ p=0.000; $q_6$ p=0.006.}
    \label{fig:know_know_llama_combined}
\end{figure}

\begin{figure}[htbp]
    \centering
    \begin{subfigure}[b]{0.24\textwidth}
        \centering
        \includegraphics[width=\textwidth]{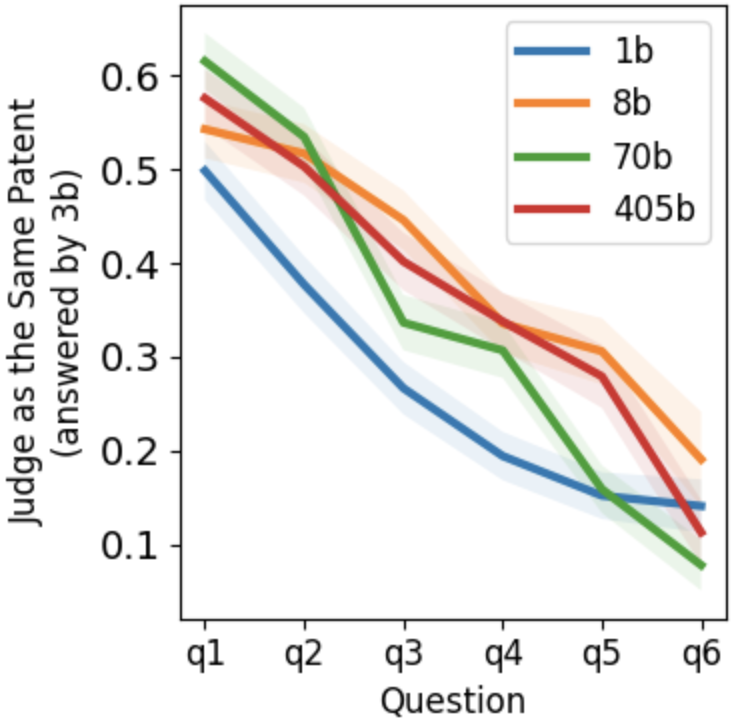}
        \label{fig:subfig1}
    \end{subfigure}
    \hfill
    \begin{subfigure}[b]{0.23\textwidth}
        \centering
        \includegraphics[width=\textwidth]{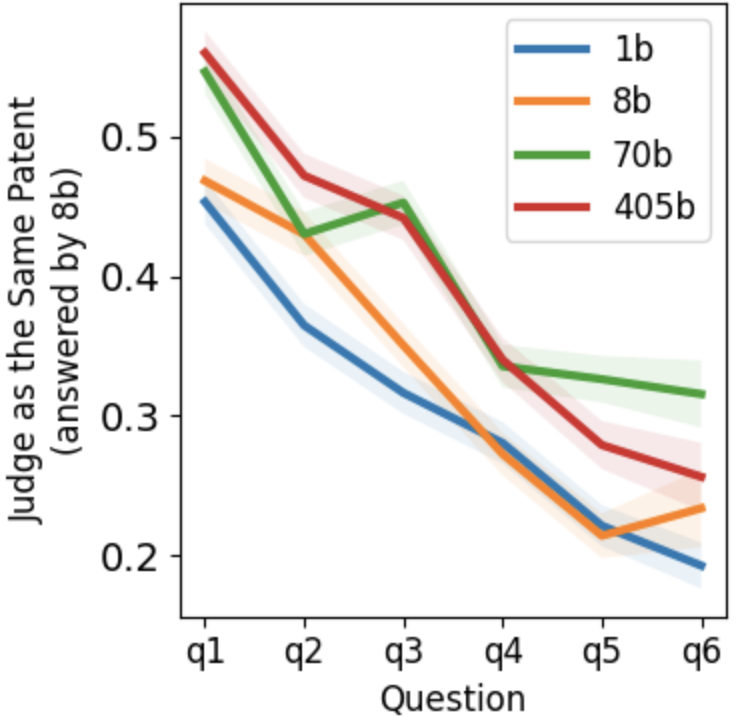}
        \label{fig:subfig2}
    \end{subfigure}
    \caption{
    \textbf{Llama 3B/8B's judge with questions proposed by different models with different sizes.} 
    Ask “Do you think they are the same patent?”, with the ground truth being *different*. We observe that bigger models' questions are less useful for the answer models to judge.
    }
    \label{fig:combined_majority}
\end{figure}
\newpage

\clearpage
\section{Pretraining Exposure}\label {pretrain}

The key to evaluating the model understanding is that the tests we created are unlikely to be in the training data. If they are, it might be that the models simply regrgitate case that they have already seen before and not demonstrate understanding. Note that we are less concerned with whether the patent texts alone are present in the pretraining data, because we deliberately construct a pairing mechanism whose likelihood of having been encountered during pretraining is extremely low. The patent set used in the paper contains roughly \(n=1.5 \times 10^{6}\) patent texts. Each text averages about \(400\) words, or approximately \(520\) byte‑pair‑encoded (BPE) tokens\footnote{Rule of thumb: one English word \(\approx 1.3\) BPE tokens. See the OpenAI Help Center article “What are tokens and how to count them?” \url{https://help.openai.com/en/articles/4936856-what-are-tokens-and-how-to-count-them}.}. The number of unordered pairs is \(\tfrac{n(n-1)}{2}\), so concatenating every possible pair would require approximately $1.2 \times 10^{15}\ \text{tokens}$. That total is multiple orders of magnitude larger than some of the largest publicly reported training corpora, even though a large share of these corpora is (unrelated) github code, books, Wikipedia, etc: GPT models were trained on \(\sim 3 \times 10^{11}\) tokens\footnote{Brown, T.B.\ \emph{et al.}(2020), “Language Models are Few‑Shot Learners.”} and Llama‑70B on \(\sim 2 \times 10^{12}\) tokens\footnote{Touvron, H.\ \emph{et al.}(2023), “Llama: Open Foundation and Fine‑Tuned LLMs.”}. The size of Redpajama English tokens is \(\sim 2 \times 10^{13}\) \footnote{\url{https://github.com/togethercomputer/RedPajama-Data}}.

\section{Example Questions}\label{qual}

\begin{figure}[h]
    \centering
    \includegraphics[width=0.48\textwidth]{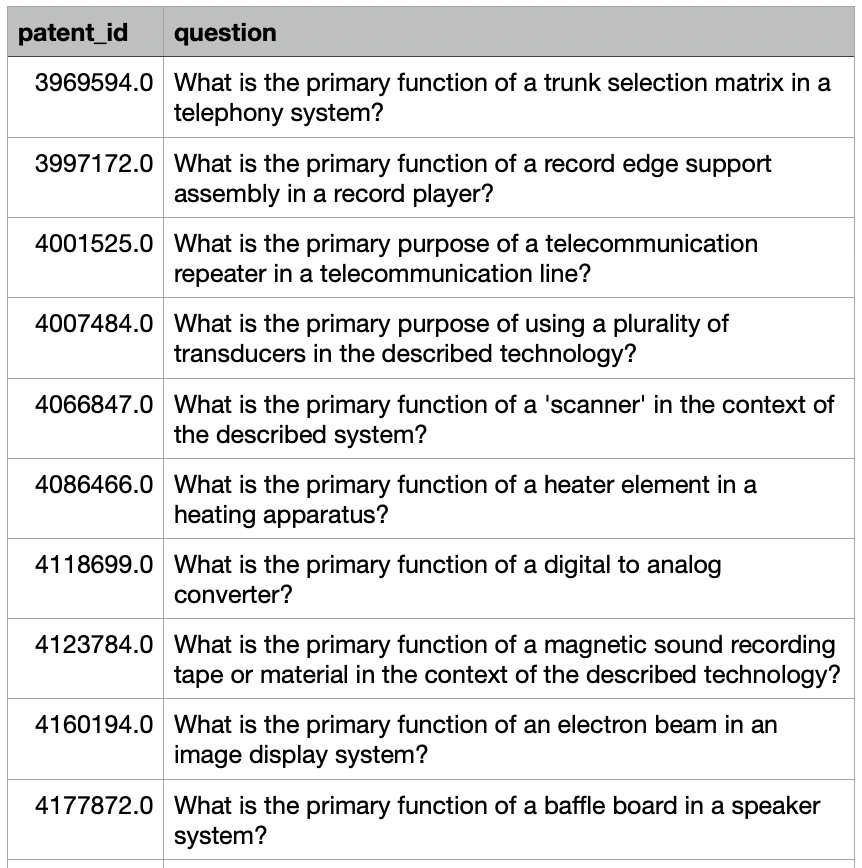} 
    \caption{Example questions.}
    \label{fig:example222}
\end{figure}

Qualitative inspection of the generated questions suggests that answering them should not require substantial novel information beyond the models’ training (echoing the missing vs. unused knowledge hypothesis). Many questions are straightforward and focus on established concepts or systems. For example, questions involving the TCP Port Service Multiplexer (TCPMUX) relate to long-standing infrastructures, with patents contributing only incremental advancements. As shown in our original paper, 68\% of the Llama 1B model’s questions and 44\% of the Llama 405B model’s questions include phrases like “what is” or “what are,” indicating the dominance of factual, definitional queries. A sample of questions can be seen in Figure \ref{fig:example222}.

\section{Ablation Studies}\label{ap:ablation}

\begin{figure}[h]
    \centering
    \includegraphics[width=0.48\textwidth]{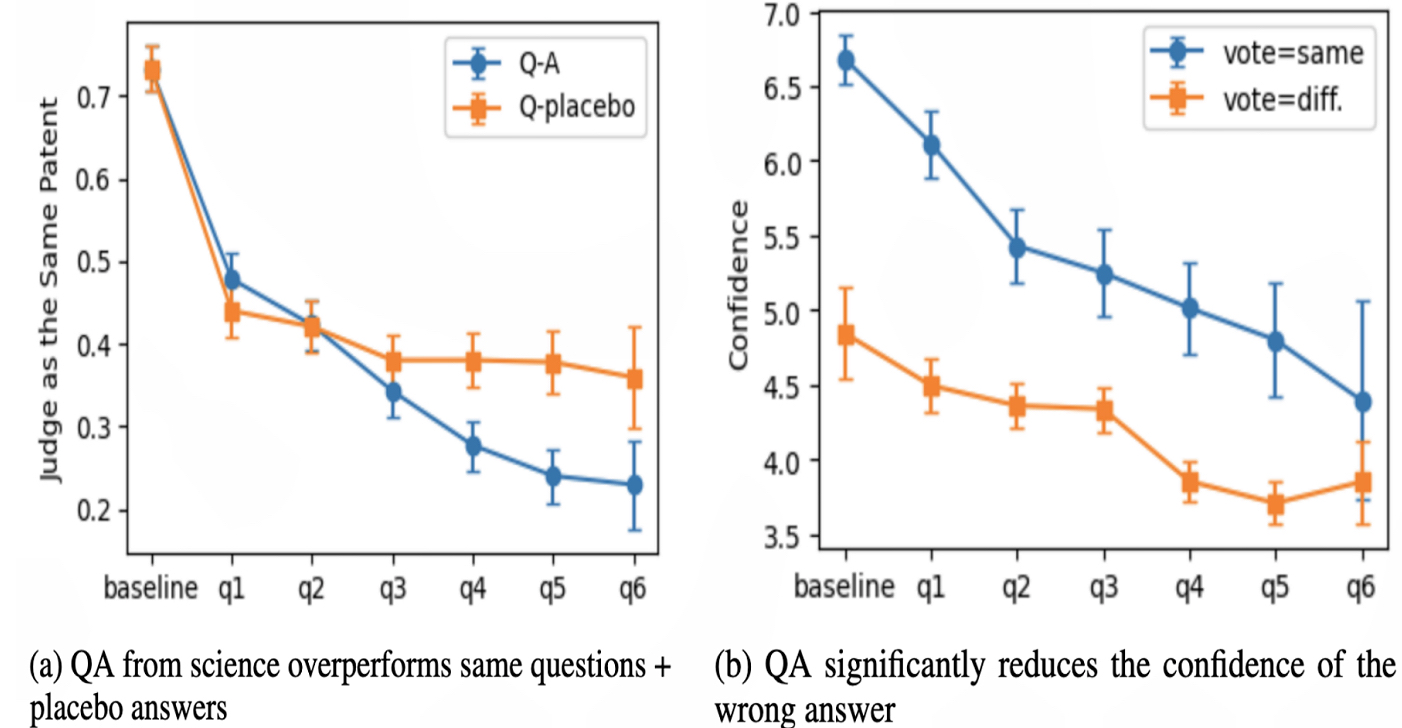} 
    \caption{More scientific-QA improves the judge over the placebo answer: For panel (a): $p_{q1}$ =0.0868, $p_{q2}$ =0.9270, $p_{q3}$ =0.0934, $p_{q4}$ =0.0000, $p_{q5}$ =0.0000, $p_{q6}$ =0.0021. We ask "are they the same?" ("yes" is the wrong answer) given the human-rated ground truth of being different patents.}
    \label{fig:example22}
\end{figure}

We gradually add scientific QA pairs from 1-6 for each patent to the original pair of patent texts, and then have Llama 3.1 8B re-judge. As shown in Figure \ref{fig:example22}, the baseline represents judge accuracy and confidence without any self-questioning or scientific retrieval. The placebo condition uses the same question but adds a randomly selected piece of text from the same paper as the found answers, matched in length to the actual answers. This is aiming for ablation studies on retrieval quality -- what if the retrieved answer is of very low quality (even unrelated)? We observe that as more QAs from science are added, judge accuracy significantly increases and becomes better than placebo answers. The confidence of both correct and incorrect answers decreases, but the drop is significantly greater for incorrect answers. Interestingly, even the placebo answer — though merely a random, same-length excerpt from the paper — enhances understanding to some extent. However, unlike actual scientific answers, its effect does not scale with the number of questions, meaning there is some initial signal that helps judgment but LLM does not learn more over more rounds of relatively unrelated science. This echoes with the findings in the main paper: (1) the question alone already facilitates understanding, functioning as cues for better deployment of internal knowledge (we explored this further in Section \ref{result2}), and (2) the placebo text still contains scientific content, which can contribute to comprehension.

\end{document}